\documentclass{article}

\usepackage{arxiv}

\usepackage[utf8]{inputenc} 
\usepackage[T1]{fontenc}    
\usepackage{hyperref}       
\usepackage{url}            
\usepackage{booktabs}       
\usepackage{amsfonts}       
\usepackage{nicefrac}       
\usepackage{microtype}      
\usepackage{lipsum}
\usepackage{graphicx}
\usepackage{enumitem}
\usepackage{tabularx}
\graphicspath{ {./images/} }

\title{Enabling Generic Robot Skill Implementation Using Object Oriented Programming}

\author{
 Abdullah Farrukh \\
  Innovative Fabric Systems\\
  German Research Institute for Artificial Intelligence\\
  Trippstadterstrasse 122, Kaiserslautern 67663 \\
  \texttt{abdullah.farrukh@dfki.de} \\
   \And
 Achim Wagner \\
  Innovative Fabric Systems\\
  German Research Institute for Artificial Intelligence\\
  Trippstadterstrasse 122, Kaiserslautern 67663 \\
  \texttt{achim.wagner@dfki.de} \\
  \And
 Martin Ruskowski \\
  Innovative Fabric Systems\\
  German Research Institute for Artificial Intelligence\\
  Trippstadterstrasse 122, Kaiserslautern 67663 \\
  \texttt{martin.ruskowski@dfki.de} \\
}

\begin{document}
\maketitle
\begin{abstract}
    Enabling Generic Robot Skill Implementation Using Object Oriented Programming
\end{abstract}


\section{Introduction}
    Developing robotic algorithms and integrating a robotic subsystem into a larger system can be a difficult task. Particularly in small and medium-sized enterprises (SMEs) where robotics expertise is lacking, implementing, maintaining and developing robotic systems can be a challenge. As a result, many companies rely on external expertise through system integrators, which, in some cases, can lead to vendor lock-in and external dependency. In the academic research on intelligent manufacturing systems, robots play a critical role in the design of robust autonomous systems. Similar challenges are faced by researchers who want to use robotic systems as a component in a larger smart system, without having to deal with the complexity and vastness of the robot interfaces in detail. In this paper, we propose a software framework that reduces the effort required to deploy a working robotic system. The focus is solely on providing a concept for simplifying the different interfaces of a modern robot system and using an abstraction layer for different manufacturers and models. The Python programming language is used to implement a prototype of the concept. The target system is a bin-picking cell containing a Yaskawa Motoman GP4.

\section{Related Work}
    Similar approaches have been pursued, implemented and adopted successfully in the past. The Robot Operating System (ROS) \cite{quigley_ros_2009} and its real-time successor ROS2 \cite{macenski_robot_2022} are widely known in the robotics community as the go-to frameworks for developing robot applications, especially in the research domain. Although the use of ROS has increased the modularity and usability of robot software, it still requires framework-specific knowledge to enable the user to develop custom applications. The Robotics Library is a C++ framework that uses an objectoriented software architecture to enable the development of robot applications \cite{rickert_robotics_2017}. The authors provide a complete set of tools in a platform-independent library, ranging from numerical methods to path planning. The authors provide a custom implementation of the hardware abstraction layer for robots as well as other devices such as sensors, grippers and more. The Open Robotics and Animation Virtual Environment (OpenRAVE) is another powerful framework for developing robotic applications \cite{diankov_openrave_2008}. Similar to ROS, OpenRAVE has a modular structure and is extensible and can be used together with ROS and Player \cite{collett_player_2005}, another robot software framework. At the time of writing, OpenRAVE provides a C++ and a Python API. In industry, the Standard Robot Command Interface (SRCI) is an emerging framework for controlling industrial robots using programmable logic controllers (PLCs) \cite{profibus_srci_nodate}. Based on a server-client architecture, the aim is to enable hardware-agnostic application design, similar to the approaches mentioned above. The PLCOpen open standard is another approachto create reusable application code in the industries \cite{eldijk_motion_2018}. Our approach focuses on non-domain users and uses pre-existing hardware abstractions to build an application layer on top of them. In addition, the aforementioned approaches do not fully consider the current and future requirements in the field of Industry 4.0 that enable a fully autonomous robotic system, which we aim to address.

    \label{chapter2}

\section{Methodology}
    In this work, we aim to create a framework to reduce the time it takes to deploy robotic applications in research. To do this, we first collect requirements for the various robotic systems used in our research department. This includes information about robot manufacturers and models, required control interfaces and middleware, and types of applications, e.g. pick and place tasks. Additional requirements are added to comply with Industry 4.0 concepts such as the Asset Administration Shell (AAS) and a standardised control interface using Open Platform Communications Unified Architecture (OPC UA) \cite{heidel_industrie_2019} \cite{opc_foundation_opc_nodate}. AAS technology is used to describe the configuration of the robot system in a standardised and interoperable way. This configuration is used as input to automate the setup of the robot system. After collecting the requirements, a software architecture is modelled using a class diagram and the basic rules and concepts of object-oriented programming are applied \cite{gamma_design_1994} \cite{poetzsch-heffter_konzepte_2009}. The modelled software architectureis then implemented using the Python programming language and tested on a bin-picking cell. As robotic systems can take different forms, this paper focuses only on 6 degrees of freedom (DOF) industrial robot arms.

\section{Requirements}
    The requirements were gathered by analysing existing software in code repositories and conducting interviews with users who have worked with industrial robots in the past, are currently developing robotic applications, or will be using robots in upcoming projects. Domain experts in the field of Industry 4.0 were asked about the requirements for a standardised control interface. In summary, the requirements collected can be grouped into three main categories: usability, compatibility and interoperability:
    \begin{enumerate}[label=\roman*)]
        \item Usability \\
            Usability requirements address the variance in users’ robot domain knowledge. The internal research showed that the majority of robot users do not have in-depth knowledge of robotic systems and only use them as a subsystem or component to achieve a higher level of automation. Therefore, the framework is required to have a user-friendly API to design the user application. It was assumed that all users of this framework would have acquired general safety instructions for the use of a robotic system.
            
        \item Compatibility \\
            Ensuring compatibility with other software modules is a key requirement of the software framework. This functionality allows the robot system to be integrated into a larger, more complex automation system that uses autonomous control of hardware modules through the use of multi-agent systems (MAS). At the same time, there are requirements for high reliability and availability of the control API. In order to get a better grip on the high variance of interfaces and middleware for different manufacturers and models, the reuse of pre-existing, manufacturer-supplied interfaces should be used in a modular way. Existing applications in the software repositories mainly used the Robot Operating System (ROS) framework. Other applications used proprietary interfaces, such as Universal Robots’ Real-Time Data Exchange (RTDE). Current and future projects already use or plan to use OPC UA interfaces, which need to be integrated into the interface layer. In addition, each manufacturer and the interfaces available for its robot must be integrated in a modular and extensible way as a basic requirement for the software architecture. As simulations play a key role in the development of robotic applications, support for simulation frameworks must be considered in the design of the framework.
            
        \item Interoperability \\
            With the aim of enabling non-experts to design robotic applications or to integrate robotic systems as a subsystem, interoperability was extracted as a key requirement. To enable this, the software framework would require a hardware and, if hardware or software constraints allow, interface-agnostic     implementation of a basic set of robot skills, with methods for adding additional, application-specific skills. If the robot application is to be used as a standalone component of a system, the external control API must ensure a widely adopted standard. In the case of integration as a component in anapplication, e.g. OPC UA server, the requirements of ”good” software engineering apply. In addition, the configuration of a robotic system must be described using state-of-the-art semantic technology AAS to ensure a common understanding of the system. This includes basic information about the robot (manufacturer, model, etc.), its interfaces (e.g. ROS, OPC UA, etc.) and interface-specific information, e.g. topic names, available services, etc., which is machine and human-readable.    
    \end{enumerate}

\section{Proposed Concept of Framework}
    \begin{figure*}[!ht]
        \centering
        \includegraphics[width=0.75\textwidth]{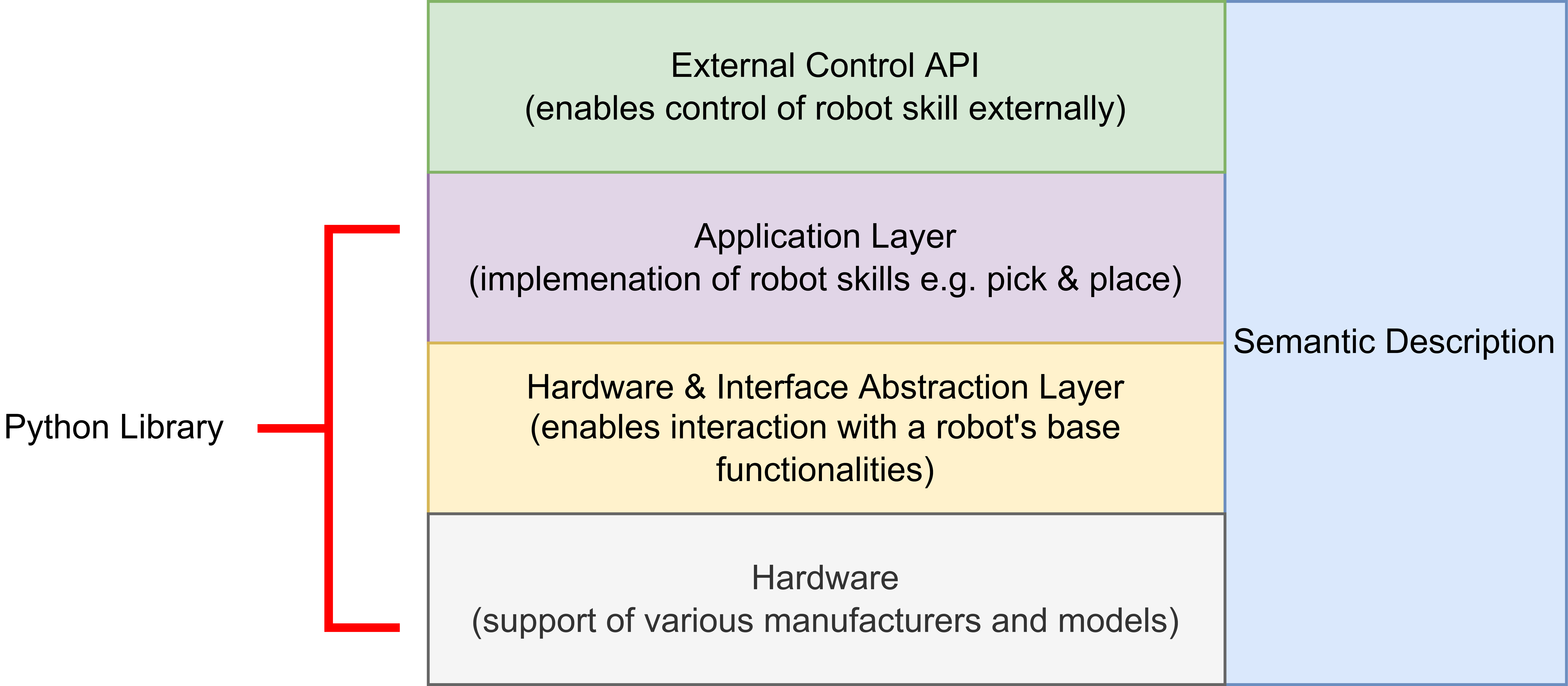}
        \caption{The proposed software framework consists of a multi-layer architecture. The lowest layer is the supported target hardware. For this, only hardware that already has hardware and interface abstractions provided by the manufacturer is chosen. The various hardware and interface abstractions are unified to enable the application layer as a Python library. The External Control API layer provides basic monitoring and error handling functionality. All layers are represented in a semantic description..}
        \label{fig:concept}
    \end{figure*}

    Considering the aforementioned requirements and existing methods, a multilayer architecture seems to be the most promising approach. By encapsulating the hardware and its abstractions from the application and control layers, a high reusability of the designed algorithms is guaranteed. In addition, the hardwareagnostic implementation of robot capabilities is enabled, which further improves the reusability of robot software and reduces deployment effort. The hardware and interface abstraction layer in our approach uses pre-existing software to control the robot system and access its parameters. These implementations are maintained by the vendor, ensuring compatibility and long-term support. For the robotic systems in our research department, ROS support is available in all cases. In addition, native interfaces and OPC UA interfaces are available for most industrial robots. The challenge is to adapt the control API of the Python library to the different manufacturers and available abstractions, which is done using Python’s extensible module design to ensure that the applications in the application layer can run independently of the hardware.

\subsection{Software Architecture \& Implementation}
    \begin{figure*} [!ht]
        \centering
        \includegraphics[width=1\textwidth]{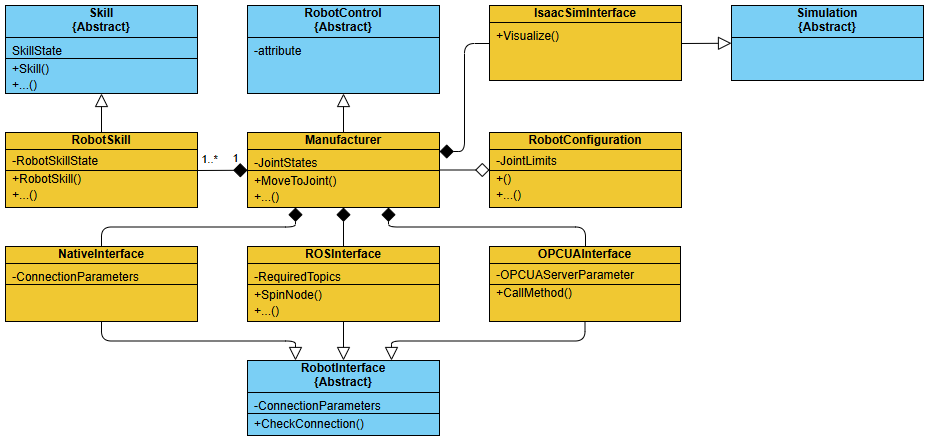}
        \caption{The class diagram represents the software architecture, which is implemented using the Python programming language. The architecture is modular in the sense that the library can be extended and maintained independently for each manufacturer. The mapping between the existing hardware abstraction (e.g. ROS) and the control API of the library is implemented in the manufacturer class. The robot capabilities are implemented by hardware independent methods defined by the abstract class RobotControl.}
        \label{fig:oop}
    \end{figure*}

    To implement the concept using object-oriented programming, a class diagram was designed for an initial prototype for the Python library, as shown in \ref{fig:oop}. Abstract classes are used to provide a design specification for each class, enforcing both consistency and flexibility for specific implementations through inheritance. The RobotControl class is the main component. It provides the structure for the control API, which is hardware independent. The Manufacturer class is used to map hardware and interface specific processes and functionalities. Composition allows multiple interfaces to be added to the Manufacturer class, with only one interface active at runtime. Each manufacturer can provide multiple interfaces to control the robot system. The abstract class RobotInterface is used to implement the different interfaces and map the control API for the target robot system to the manufacturer class. Since skills can be of different scopes and contain different functionality, the abstract Skill class provides the basic structure for the RobotSkill class. Each defined robot skill implements the RobotSkill class using the methods of the RobotControl class, allowing hardware-agnostic algorithm development. In our work we have also placed an abstract class for simulation to cover some aspects of simulations. Currently, only a lightweight Isaac SimInterface class uses this class to enable visualisation.
    
\subsection{Results \& Discussion}
    The prototype of the Python library was used to replace the control software of a bin-picking cell, which used software with a near-monolithic software architecture, with just a single class to allow control via the ROS API. The existing software was written specifically for the cell, although a ROS interface was used to control the robotic system. The metric of lines of code (LOC) was introduced to compare the efficiency of our approach with conventional methods. The data is only for a singular use-case and results may vary in other use-cases.

    \begin{table}[!ht]
        \begin{center}
            \begin{tabular}{l>{\centering\arraybackslash}p{3.5cm}>{\centering\arraybackslash}p{3.5cm}}
                \hline
                \multicolumn{1}{l}{\rule{0pt}{12pt} Software Architecture} 
                & \multicolumn{1}{c}{LoC\footnotemark[1] - specialized code} 
                & \multicolumn{1}{c}{LoC - reusable code} \\[2pt]
                \hline
                \rule{0pt}{12pt}
                Monolithic      & 56    & 212  \\
                Our Approach    & 23    & 434 \\[2pt]
                \hline \\[0.75pt]
            \end{tabular} \\
        \hspace*{-\leftskip}
        \footnotemark[1]{Lines of Code}
        \caption{THE TABLE COMPARES THE NUMBER OF LINES OF CODE USED TO IMPLEMENT A BIN-PICKING ALGORITHM AS A METRIC TO HIGHLIGHT THE ADVANTAGE OF OUR APPROACH OVER A CONVENTIONAL, MONOLITHIC APPROACH TO PROGRAMMING ROBOT APPLICATIONS.}
        \label{table:LoC}
        \end{center}
    \end{table}

    Table \ref{table:LoC} summarises the number of lines of code required to implement the bin-picking algorithm. The old monolithic architecture used ROS2 as middleware to control the robot system and was incompatible with other hardware abstractions.This resulted in the specialised code also having ROS2 functionalities, making the algorithm unusable for other interfaces. Although our approach has more lines of reusable code, it also allows the use of alternative interfaces such as OPC UA. Also, the specialised code does not use any hardware or interface specific functionality, making the algorithm usable with other robot systems and interfaces. In summary, the proposed framework separates the complexity of the hardware and interfaces of robotic systems from the application layer to enable users with little knowledge of the robotics domain to develop and deploy robotic applications. Users are still required to have a basic knowledge of robotic systems, particularly in the area of safety. The modular design of the developed Python library requires the integration of vendor specific implementations as separate packages, which reduces maintenance effort and provides a clean software structure. The implementation of a basic set of robot skills was a challenge, as there is no common understanding or definition that could have been used, other than the industrial norms and standards mentioned in chapter \ref{chapter2}. Therefore, future work needs to focus on this area. In addition, the performance in real-time dynamic scenarios needs to be evaluated. In addition, as a prerequisite for using this library, the environment for the specific interface must be set, e.g. ROS2 drivers.

\section{Conclusion}
    The proposed software framework increases the efficiency of robotic application development. The most practical aspect is the complete hardware and interface independent development of the application algorithm. In the aforementioned bin-picking cell use case, the code developed can be reused for any robot manufacturer or model integrated into the framework by simply adapting the configuration. Although the implementation in this paper was prototypical, no major errors were detected during runtime for the given scenario. This is not to say that rigorous software testing is not required to confirm this aspect. In future work, support for the simulation framework needs to be addressed. Also, the definition of robot capabilities needs to be based on a thorough literature review.

    This research has been supported by the European Union’s HORIZON Research and Innovation Action Program under the grant agreement No 101138782, the project \textbf{RAASCEMAN}\footnote{https://cordis.europa.eu/project/id/101138782} and by the German Federal Ministry for Economic Affairs and Climate Action (BMWK) in the context of \textbf{TWIN4TRUCKS}\footnote{https://www.twin4trucks.de/} project (13IK010F).

\bibliographystyle{unsrt}  

\bibliography{raad2025}

\end{document}